\long\def\plcmdam{$\avm{ \mbox{\sc verbal}\\
\mbox{\it sem}: \mbox{want(Subj,able(Subj,give(Subj,books,arie)))}}
$
}
\long\def\plcmdan{$
\xavm[{\mbox{J}}]{ \mbox{\sc noun}  \\
 \mbox{\it lex} : \mbox{arie}  \\
 \mbox{\it dir} : \mbox{left} }$
}
\long\def\plcmdao{$
\xavm[{\mbox{L}}]{ \mbox{\sc noun}  \\
 \mbox{\it lex} : \mbox{boeken}  \\
 \mbox{\it dir} : \mbox{left} }$
}
\long\def\plcmdap{$
\avm{
 \mbox{\sc verbal}  \\
 \mbox{\it lex} : \mbox{wil}  \\
 \mbox{\it sc} :\langle \mbox{P} , \mbox{R} , \mbox{L}, \mbox{J} \rangle}$
}
\long\def\plcmdaq{$\xavm[{\mbox{P}}]{ \mbox{\sc verbal}  \\
 \mbox{\it lex} : \mbox{kunnen}  \\
 \mbox{\it sc} :\langle \mbox{R}, \mbox{L}, \mbox{J} \rangle  \\
 \mbox{\it dir} : \mbox{right} }
$
}
\long\def\plcmdar{$\xavm[{\mbox{R}}]{ \mbox{\sc verbal}  \\
 \mbox{\it lex} : \mbox{geven}  \\
 \mbox{\it sc} : \langle \mbox{L},\mbox{J} \rangle  \\
 \mbox{\it dir} : \mbox{right} }
$
}

\documentstyle[a4wide,fullname,fleqn]{article}

\long\def\xavm[#1]#2{\left[ \begin{array}{l}#2 %
                \end{array} \right]_{#1} \vspace{0.5ex}}

\long\def\avm{\xavm[{}]}

\newcounter{sub}[equation]

\def\theexs {\arabic{equation}}

\newenvironment{exam}{\refstepcounter{equation}\begin{list}{(\theexs)}\protect\item}{\end{list}}

\catcode`\@=11

\newif\ifLaTeX                            
\ifx\@ndefined\@@startpbox\LaTeXfalse     
    \else\LaTeXtrue\fi                    
\immediate\write16{This is `Treemaker' Version 1.232 for use with \ifLaTeX
LaTeX%
                   \else plain TeX\fi}

\ifLaTeX               
   \else \let\@line\line                
         \input l_pic                   
         \let\line\@line                
   \fi                                  

\catcode`\@=11
\let\g\global
\def\gxdef{\global\xdef}

\def\newcount{\alloc@0\count\countdef\insc@unt}

\def\for#1:=#2\to#3\do#4\od{%
   \def\f@rcount{#1}\def\upp@rlimit{#3}\def\b@dy{#4}\f@rcount=#2\relax\dof@r}

\def\dof@r{\ifnum\f@rcount>\upp@rlimit\relax\let\n@xt\relax
                 \else\b@dy\advance\f@rcount\@ne\let\n@xt\dof@r\fi
           \n@xt}

\newcount\@xcount
\newcount\t@mes

\def\ex#1\times#2\xe{%
   \@xcount1 \t@mes#1\def\b@dy{#2}\do@x}

\def\do@x{\ifnum\@xcount>\t@mes\let\n@xt\relax
                \else\b@dy\advance\@xcount\@ne\let\n@xt\do@x\fi
          \n@xt}

\newskip\thickn@ss
\newskip\@nner
\newskip\@uter

\def\rect@ngle#1#2#3{\hbox to 0pt{%
     \thickn@ss#3%
     \g\@nner#2\g\advance\@nner-\thickn@ss
     \g\divide\@nner\tw@
     \g\@uter#2\g\advance\@uter\thickn@ss
     \g\divide\@uter\tw@
     \hskip 0pt minus .5fil%
     \vrule height\@uter depth\@nner width\thickn@ss
     \vrule height\@uter depth-\@nner width#1%
     \hskip 0pt minus 1fil%
     \vrule height-\@nner depth\@uter width#1%
     \vrule height\@nner depth\@uter width\thickn@ss
     \hskip 0pt minus .5fil%
     }
     }

\def\s@ries#1#2{%
     \g\t@mpcnta1
     \gdef\t@mp{#1}%
     \@ssign#2/\l@st  
     \expandafter\gdef\csname#1\endcsname##1{%
                      \csname#1\romannumeral##1\endcsname}%
     }

\def\@ssign#1/#2{%
      \expandafter\gdef\csname\t@mp\romannumeral\t@mpcnta\endcsname{#1}%
      \g\advance\t@mpcnta\@ne
      \ifx#2\l@st
          \g\let\n@xt\relax
     \else\g\let\n@xt\@ssign
       \fi
      \n@xt}

\newdimen\leftdist
\newdimen\rightdist
\newbox\TeXTree

\newcount\sl@pe
\newcount\l@vels
\newcount\s@ze

\newbox\circleb@x
\newbox\squareb@x
\newbox\dotb@x
\newbox\triangleb@x
\newbox\textb@x
\newbox\frameb@x

\newdimen\circlew@dth
\newdimen\squarew@dth
\newdimen\dotw@dth
\newdimen\trianglew@dth
\newdimen\textw@dth
\newdimen\framew@dth

\newdimen\vd@st
\newdimen\hd@st
\newdimen\based@st
\newdimen\dummyhalfcenterdim@n

\newcount\t@mpcnta
\newcount\t@mpcntb
\newcount\t@mpcntc
\newcount\t@mpcntd
\newdimen\t@mpdima
\newdimen\t@mpdimb
\newdimen\t@mpdimc
\newbox\t@mpboxa
\newbox\t@mpboxb

\newbox\leftb@x
\newbox\rightb@x
\newbox\centerb@x
\newbox\beneathb@x
\newtoks\typ@
\newbox\centerb@@x
\newdimen\centerdim@n
\newdimen\halfcenterdim@n

\s@ries{xv@l}{0//1//1//1//1//2//1//3//2//3//4//5//1//6//%
              5//4//3//5//2//5//3//4//5//6}
\s@ries{yv@l}{1//6//5//4//3//5//2//5//3//4//5//6//1//5//%
              4//3//2//3//1//2//1//1//1//1}

\def\hv@ldef{%
     \for\t@mpcnta:=1\to24%
      \do\g\t@mpdima\vd@st\g\multiply\t@mpdima by\xv@l{\t@mpcnta}%
         \g\divide\t@mpdima by\yv@l{\t@mpcnta}\g\multiply\t@mpdima by 2
         \expandafter\gxdef\csname hv@l\romannumeral\t@mpcnta\endcsname{%
                          \the\t@mpdima}%
      \od}

\def\hv@l#1{\csname hv@l\romannumeral#1\endcsname}

\newcount\depth
\newdimen\help
\newif\ifreflect
\newif\ifalfl
\newif\ifblfl
\newif\ifrlfl
\newif\ifllfl
\newif\ifedfl
\newif\ifclfl
\newif\iftemp

\gdef\rotation#1
{ \ifnum#1=90 \immediate\write-1{> rotation #1}\fi
  \ifnum#1=180 \immediate\write-1{> rotation #1}\fi
}

\gdef\pictexmode{\immediate\write-1{> pictexmode}}

\gdef\expand#1{\expandafter #1}

\gdef\tree#1{
\ifclfl
  \errmessage{running over end of treedef, possibly you said somewhere `leaf{}'
instead of `tree{}'}
\else
  \alflfalse
  \llflfalse
  \rlflfalse
  \edflfalse
  \blflfalse
  \setbox\depth=\hbox{\expand #1}\batchmode
  \showbox\depth%
  \errorstopmode
  \immediate\write-1{> node #1&}
  \beginsubtree
\fi
}

\gdef\andtree#1{
\ifclfl
  \errmessage{running over end of treedef, possibly you said somewhere `leaf{}'
instead of `tree{}'}
\else
  \alflfalse
  \llflfalse
  \rlflfalse
  \edflfalse
  \blflfalse
  \setbox\depth=\hbox{\expand #1}\batchmode
  \showbox\depth%
  \errorstopmode
  \immediate\write-1{> anode #1&}
  \beginsubtree
\fi
}

\gdef\leaf#1{
\alflfalse
\llflfalse
\rlflfalse
\edflfalse
\blflfalse
\setbox\depth=\hbox{\expand #1}\batchmode
\showbox\depth%
\errorstopmode
\immediate\write-1{> leaf #1&}
}

\gdef\subtree#1{
\temptrue
\ifrlfl
  \errmessage{sorry, but you can't use `subtree' together with `rlabel'}
  \tempfalse
\fi
\ifllfl
  \errmessage{sorry, but you can't use `subtree' together with `llabel'}
  \tempfalse
\fi
\ifalfl
  \errmessage{sorry, but you can't use `subtree' together with `alabel'}
  \tempfalse
\fi
\iftemp
  \tempfalse
  \blflfalse
  \edflfalse
  \setbox\depth=\hbox{}\batchmode
  \showbox\depth%
  \errorstopmode
  depth \number\depth\ subtree: \box\depth\ \par
  \immediate\write-1{> subtree }
  \beginsubtree
  \ifreflect
    \leaf{}
    \llabel{#1}
    \leaf{}
  \else
    \rlabel{#1}
    \leaf{}
    \leaf{}
  \fi
  \endtree
\fi
}

\gdef\edge#1{
\ifedfl
  \errmessage{too many `edge' detected}
\else
  \edfltrue
  \setbox\depth=\hbox{\expand #1}\batchmode
  \showbox\depth%
  \errorstopmode
  edge: \box\depth\par \immediate\write-1{> edge #1&}
\fi
}

\gdef\mboxedge#1{
\ifedfl
  \errmessage{too many `edge' detected}
\else
  \edfltrue
  \setbox\depth=\hbox{\expand #1}\batchmode
  \showbox\depth%
  \errorstopmode
  edge: \box\depth\par \immediate\write-1{> mboxedge #1&}
\fi
}

\gdef\llabel#1{
\temptrue
\ifllfl
  \errmessage{too many `llabel' detected}
  \tempfalse
\fi
\ifalfl
  \errmessage{you can not use `alabel' together with `llabel}
  \tempfalse
\fi
\iftemp
  \tempfalse
  \llfltrue
  \setbox\depth=\hbox{\expand #1}\batchmode
  \showbox\depth%
  \errorstopmode
  \ifreflect
    rlabel: \box\depth\par \immediate\write-1{> rlabel #1&}
  \else
    llabel: \box\depth\par \immediate\write-1{> llabel #1&}
  \fi
\fi
}

\gdef\rlabel#1{
\temptrue
\ifrlfl
  \errmessage{too many `rlabel' detected}
  \tempfalse
\fi
\ifalfl
  \errmessage{you must not use `alabel' together with `rlabel'}
\tempfalse
\fi
\iftemp
  \tempfalse
  \rlfltrue
  \setbox\depth=\hbox{\expand #1}\batchmode
  \showbox\depth%
  \errorstopmode
  \ifreflect
\immediate\write-1{> llabel #1&}
  \else
\immediate\write-1{> rlabel #1&}
  \fi
\fi
}

\gdef\blabel#1{
\ifblfl
  \errmessage{too many `blabel' detected}
\else
  \blfltrue
  \setbox\depth=\hbox{\expand #1}\batchmode
  \showbox\depth%
  \errorstopmode
  blabel: \box\depth\par \immediate\write-1{> blabel #1&}
\fi
}

\gdef\alabel#1{
\temptrue
\ifalfl
  \errmessage{too many `alabel' detected}
  \tempfalse
\fi
\ifllfl
  \errmessage{you can not use `alabel' together with `llabel'}
  \tempfalse
\fi
\ifrlfl
  \errmessage{you must not use `alabel' together with `rlabel'}
  \tempfalse
\fi
\iftemp
  \alfltrue
  \tempfalse
  \setbox\depth=\hbox{#1}\batchmode
  \showbox\depth%
  \errorstopmode
  alabel: \box\depth\par \immediate\write-1{> alabel #1&}
\fi
}

\gdef\opentree#1{
\reflectfalse
\alflfalse
\blflfalse
\rlflfalse
\llflfalse
\edflfalse
\clflfalse
\tempfalse
\ifx\@ndefined\plotsymbolspacing  
  \else\pictexmode
\fi
  \depth=0
  \immediate\write-1{> beginTree #1&}
}

\gdef\closetree{\immediate\write-1{> endTree}
}

\gdef\beginsubtree{\advance\depth by1
}

\gdef\endtree{\advance\depth by-1
  \ifnum\depth<0 \errmessage{too many `endtree' detected} \fi
  \ifnum\depth=0
    \closetree
    \clfltrue
   \fi
}

\begin{document}
\bibliographystyle{fullname}

\vspace{0.2em}

\begin{center}
\LARGE Adjuncts and the Processing of Lexical Rules
\end{center}

\begin{center}
\Large Gertjan van Noord and Gosse Bouma
\end{center}

\begin{center}
BCN RUG Groningen\\
\tt \{vannoord,gosse\}@let.rug.nl
\end{center}

\vspace{2em}

\begin{abstract}
The standard HPSG analysis of Germanic verb clusters can not explain
the observed narrow-scope readings of adjuncts in such verb clusters.

We present an extension of the HPSG analysis that accounts for the
systematic ambiguity of the scope of adjuncts in verb cluster
constructions, by treating adjuncts as members of the subcat list.
The extension uses powerful recursive lexical rules,
implemented as complex constraints. We show how `delayed evaluation'
techniques from constraint-logic programming can be used to process
such lexical rules.\\

\end{abstract}

\section{Problem Description}

\subsection{Dutch Verb Clusters}
Consider the following Dutch subordinate sentences.

\begin{exam}
\label{one}
\begin{flushleft}
dat Arie wil slapen \\
that Arie wants to-sleep
\end{flushleft}
\end{exam}\begin{exam}
\begin{flushleft}
dat Arie Bob wil slaan \\
that Arie Bob wants to-hit \\
that Arie wants to hit Bob
\end{flushleft}
\end{exam}\begin{exam}
\label{three}
\begin{flushleft}
* dat Arie Bob wil slapen\\
* that Arie Bob wants to-sleep\\
* that Arie wants to sleep Bob
\end{flushleft}
\end{exam}

\begin{exam}
\label{four}
* dat Arie wil Bob slaan
\end{exam}\begin{exam}
\begin{flushleft}
 dat Arie Bob cadeautjes wil geven \\
that Arie Bob presents want to-give \\
that Arie wants to give presents to Bob
\end{flushleft}
\end{exam}\begin{exam}\label{six}
\begin{flushleft}
* dat Arie Bob wil cadeautjes geven\\
* dat Arie wil Bob cadeautjes geven
\end{flushleft}
\end{exam}

\begin{exam}
\label{seven}
\begin{flushleft}
dat Arie Bob zou moeten kunnen willen kussen \\
that Arie Bob should must can want to-kiss \\
that Arie should be able to want to kiss Bob
\end{flushleft}
\end{exam}

The examples~\ref{one}-\ref{three} indicate that in Dutch the
arguments of a main verb can be realized to the left of an intervening
auxiliary verb, such as a modal verb. Furthermore the sentences
in~\ref{four}-\ref{six} indicate that in such constructions the arguments must
be
realized to the left of the auxiliary verbs. In~\ref{seven} it is
illustrated that there can be any number of auxiliaries.

\subsection{The HPSG analysis of verb-clusters}

The now standard analysis within HPSG of such verb-clusters is based
on ideas from Categorial Grammar (cf. for example
\cite{moortgat-diss}) and defined within the HPSG
framework by \cite{hinrichs:89}. In this analysis auxiliary verbs
subcategorize for an unsaturated verb-phrase and for the
complements that are not yet realized by this verb-phrase. In other
words, the arguments of the embedded verb-phrase are inherited by the
auxiliary.

\begin{figure}
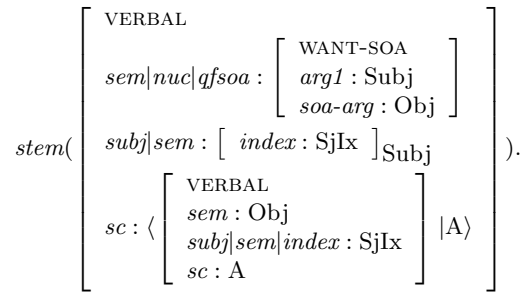

\small
\begin{center}$
\mbox{\it stem} (\avm{
 \mbox{\sc verbal}  \\
 \mbox{\it sem$|$nuc$|$qfsoa} : \avm{\mbox{\sc want-soa}\\
                                     \mbox{\it arg1} : \mbox{Subj}\\
	                             \mbox{\it soa-arg} : \mbox{Obj}}\\
 \mbox{\it subj$|$sem} : \xavm[\mbox{Subj}]{\mbox{\it index}: \mbox{SjIx}}\\
 \mbox{\it sc} :\langle \avm{
 \mbox{\sc verbal}  \\
 \mbox{\it sem} : \mbox{Obj}\\
 \mbox{\it subj$|$sem$|$index} : \mbox{SjIx}\\
 \mbox{\it sc} :\mbox{A} } | \mbox{A}\rangle }).
$\end{center}
\caption{\label{wil}The modal auxiliary `wil'.}
\end{figure}

\begin{figure*}[t,b]
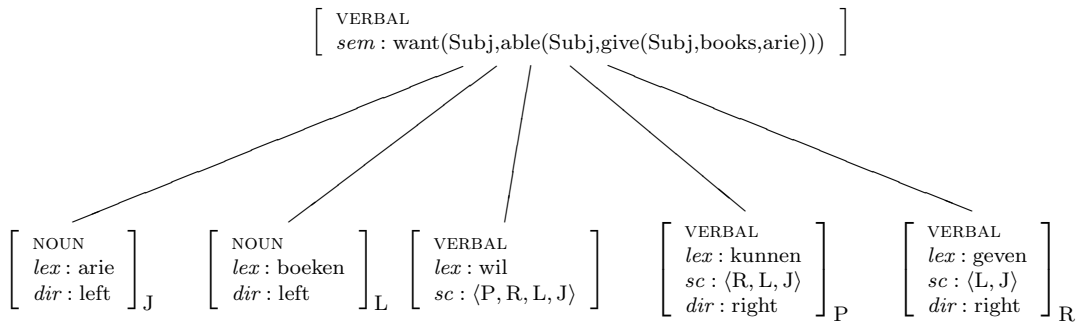

\footnotesize
\leavevmode
\unitlength1pt
\picture(406.70,137.00)
\catcode`\@=11
 \put(115.85,108.00){\hbox{\plcmdam }}
 \put(201.15,94.00){\hbox{}}
 \put(173.6533,97.0000){\line(-5,-2){147.4997}}
 \put(0.00,18.00){\hbox{\plcmdan }}
 \put(30.13,-1.50){\hbox{}}
 \put(186.4866,97.0000){\line(-4,-3){78.6665}}
 \put(75.25,18.00){\hbox{\plcmdao }}
 \put(105.96,-1.50){\hbox{}}
 \put(199.3199,97.0000){\line(-1,-6){9.8333}}
 \put(151.68,18.00){\hbox{\plcmdap }}
 \put(191.28,-1.50){\hbox{}}
 \put(214.3532,97.0000){\line(6,-5){66.0001}}
 \put(245.87,18.00){\hbox{\plcmdaq }}
 \put(284.27,-5.50){\hbox{}}
 \put(228.6532,97.0000){\line(5,-2){137.5002}}
 \put(337.66,18.00){\hbox{\plcmdar }}
 \put(372.18,-5.50){\hbox{}}
 \put(0.00,105.00){\hbox{}}
\endpicture
\normalsize
\caption{\label{deriv} The parse tree for the verb-phrase `arie boeken wil
kunnen geven'.}
\end{figure*}

For example, the auxiliary `wil' might be defined as in figure~\ref{wil}.
If we assume an application rule that produces flat vp-structures,
then we obtain the derivation in figure ~\ref{deriv} for the
infinite verb-phrase
\begin{exam}
\dots Arie boeken wil kunnen geven
\end{exam}

\subsection{Problems with the scope of adjuncts}

A major problem that this analysis faces is the possibillity of
narrow-scope readings in the case of adjuncts.
For example, the following Dutch subordinate sentences are all
systematically ambiguous between a wide-scope reading (adjunct
modifies the event introduced by the auxiliary) or a narrow-scope
reading (adjunct modifes the event introduced by the main verb).
\begin{exam}
\begin{flushleft}
dat Arie vandaag Bob wil slaan\\
that Arie today Bob want to-hit\\
that Arie wants to hit Bob today
\end{flushleft}
\end{exam}\begin{exam}
\begin{flushleft}
dat Arie het artikel op tijd probeerde op te sturen\\
that Arie the article on time tried to send\\
that Arie tried to send the article in time
\end{flushleft}
\end{exam}\begin{exam}
\begin{flushleft}
dat Arie Bob de vrouwen met een verrekijker zag bekijken\\
that Arie Bob the women with the telescope saw look-at\\
that Arie saw Bob looking at the women with the telescope
\end{flushleft}
\end{exam}

Firstly note that the treatment of adjuncts as presented in
\cite{hpsg2}, 
cannot be maintained as
it simply fails to derive any of these sentences because the
introduction of adjuncts is only possible as sisters of saturated
elements. The fact that arguments and adjuncts can come
interspersed (at least in languages such as Dutch and German) is not
accounted for.

A straightforward solution to this problem is presented in
\cite{kasper-adjuncts}. Here adjuncts and arguments are all
sisters to a head. The arguments should satisfy the subcat
requirements of this head -- the adjuncts modify the semantics of the
head (via a recursively defined adjuncts principle).

The main problem for this treatment of adjuncts is that it cannot
explain the narrow-scope readings observed above. If adjuncts modify
the head of the phrase they are part of then we will only obtain the
wide-scope readings.

If we assume, on the other hand, that adjuncts {\em are} on the subcat
list, then we will obtain both readings straightforwardly. In the
narrow-scope case the adjunct is on the subcat list of the embedded
verb, and then inherited by the matrix verb. In the wide-scope case
the adjunct simply is on the subcat list of the matrix verb. In the
next section we present a treatment of adjuncts in which each adjunct
is subcategorized for. By means of lexical rules we are able to obtain
the effect that there can be any number of adjuncts. We also sketch
how the semantics of modification might be defined.

\section{Adjuncts as Arguments}

\subsection{Adding adjuncts}

The previous section presented an argument that VP modifiers are
selected for by the verb. Note that this is in line with earlier
analyses of adjuncts in HPSG \cite{hpsg} which where abandoned as it
was unclear how the semantic contribution of adjuncts could be
defined.

Here we propose a solution in which adjuncts are members of the subcat
list, just like ordinary arguments. The difference between arguments
and adjuncts is that adjuncts are `added' to a subcat list by a
lexical rule that operates recursively.\footnote{cf. \cite{miller-diss} for a
similar suggestions concerning French.}
Such a lexical rule might for example be stated as in figure~\ref{lr}.

\begin{figure*}[t,b]
\begin{center}\small$
\avm{\mbox{\sc verbal}\\\mbox{\it sc}: \mbox{P}\cdot\mbox{S}\\
                        \mbox{\it sem}: \mbox{Sem}_0} \Longrightarrow
\avm{\mbox{\sc verbal}\\ {\it sc}:\mbox{P}\cdot\langle \avm{\mbox{\sc
adverbial}\\
                                                     \mbox{\it mod}:
\avm{\mbox{\sc mod}\\
                                     \mbox{\it arg}:\mbox{Sem}_0\\\mbox{\it
val}:\mbox{Sem} } } \rangle \cdot \mbox{S}\\
                        \mbox{\it sem}: \mbox{Sem}}
$\end{center}
\caption{\label{lr}
A lexical rule that adds a single adjunct to the subcat list of a verb. In the
case
of $n$ adjuncts the rule applies $n$ times.}
\end{figure*}

Note that in this rule the construction of the semantics of a modified
verb-phrase is
still taken care of by a {\it mod} feature on the adjunct, containing
a {\it val} and {\it arg} attribute. The {\it arg} attribute is unified with
the
`incoming' semantics of the verb-phrase without the adjunct. The {\it val}
attribute
is the resulting semantics of the verb-phrase including the adjunct.
This allows the following treatment of the semantics of modification
\footnote{inspired by \cite{kasper-adjuncts}}, cf. figure~\ref{adi}.

\begin{figure*}[t,b]
\begin{center}\small
$
\avm{\mbox{\sc restr\_adverbial}\\
     \mbox{\it mod}: \avm{ \mbox{\it arg$|$nuc} : \avm{\mbox{\it qfsoa} :
\mbox{Q}\\
                                                 \mbox{\it restr} : \mbox{R}}\\
                           \mbox{\it val$|$nuc} : \avm{\mbox{\it qfsoa} :
\mbox{Q}\\
                                                 \mbox{\it restr} : \langle
\mbox{R}_0 | \mbox{R}\rangle} } }
{}~~~ \avm{\mbox{\sc op\_adverbial}\\
     \mbox{\it mod}: \avm{ \mbox{\it arg} : \mbox{Soa}\\
                           \mbox{\it val$|$nuc} : \avm{ \mbox{\it qfsoa} :
                  \avm{\mbox{\sc accidental-soa}\\
                       \mbox{\it soa-arg}: \mbox{Soa}} \\
                                                  \mbox{\it restr} :
\langle\rangle  } }}
$
\end{center}
\caption{\label{adi}A restrictive adverbial and an operator adverbial.
Restrictive adverbials (such as locatives and time adverbials) will
generally be encoded as presented, where $\mbox{R}_0$ is a meta-variable that
is
instantiated by the restriction introduced by the adjunct.
Operator adverbials (such as causatives) on the other hand introduce their own
quantified state of affairs. Such adverbials generally are encoded as in the
following
example of the adverbial `toevallig' (accidentally).
Adverbials of the first type add a restriction to the semantics of the
verb; adverbials of the second type introduce a new scope of modification.}
\end{figure*}

We are now in a position to explain the observed ambiguity of adjuncts
in verb-cluster constructions. Cf.:
\begin{exam}
\begin{flushleft}
dat Arie Bob vandaag wil kussen\\
that Arie Bob today wants to-kiss
\end{flushleft}
\end{exam}
In the narrow-scope reading the adjunct is first added to the subcat
list of `kussen' and then passed on to the subcat list of the
auxiliary verb. In the wide-scope reading the adjunct is added to the
subcat list of the auxiliary verb. The final instantiations of the
auxiliary `wil' for both readings are given in figure~\ref{narwid}.

\begin{figure*}[t,b]

\begin{center}\small$
\avm{
 \mbox{\sc verbal}  \\
 \mbox{\it sc} :\langle \avm{
 \mbox{\sc verbal}  \\
 \mbox{\it sc} :\langle \mbox{E} , \mbox{H} \rangle  \\
 \mbox{\it lex} : \mbox{kussen}  \\
 \mbox{\it dir} : \mbox{right} } , \xavm[{\mbox{E}}]{ \mbox{\sc adverbial}  \\
 \mbox{\it lex} : \mbox{vandaag}  \\
 \mbox{\it dir} : \mbox{left} } , \xavm[{\mbox{H}}]{ \mbox{\sc noun}  \\
 \mbox{\it lex} : \mbox{bob}  \\
 \mbox{\it dir} : \mbox{left} }, \avm{
 \mbox{\sc noun}  \\
 \mbox{\it lex} : \mbox{arie}  \\
 \mbox{\it dir} : \mbox{left} }\rangle  \\
 \mbox{\it lex} : \mbox{wil} }
$ \\

$\avm{
 \mbox{\sc verbal}  \\
 \mbox{\it sc} :\langle \avm{
 \mbox{\sc verbal}  \\
 \mbox{\it sc} :\langle \mbox{H} \rangle  \\
 \mbox{\it lex} : \mbox{kussen}  \\
 \mbox{\it dir} : \mbox{right} } , \avm{
 \mbox{\sc adverbial}  \\
 \mbox{\it lex} : \mbox{vandaag}  \\
 \mbox{\it dir} : \mbox{left} } , \xavm[{\mbox{H}}]{ \mbox{\sc noun}  \\
 \mbox{\it lex} : \mbox{bob}  \\
 \mbox{\it dir} : \mbox{left} }, \avm{
 \mbox{\sc noun}  \\
 \mbox{\it lex} : \mbox{arie}  \\
 \mbox{\it dir} : \mbox{left} }\rangle  \\
 \mbox{\it lex} : \mbox{wil} } $
\end{center}

\caption{\label{narwid}The final instantiation of the modal for both
the narrow- and the wide-scope reading of the sentence `Arie Bob
vandaag wil kussen'. In the narrow-scope reading the adverbial occurs
both on the subcat list of the embedded verb and on the subcat list of
the matrix verb --- indicating that the embedded verb introduced the
adjunct. In the wide-scope reading the adverb only occurs on the
subcat list of the matrix verb.}
\end{figure*}

\subsection{Discussion}

A further problem concerning the syntax of adjuncts is posed by the
fact that adjuncts can take part in unbounded dependency
constructions.
Lexical treatments of the kind presented in
\cite[chapter 9]{hpsg2} assume that
a lexical rule is responsible for `moving' an element from the subcat
list to the slash list. Such an account predicts that adjuncts can not
take part in such unbounded dependency constructions.  In
\cite[chapter 9]{hpsg2} a special rule is introduced to account
for those cases where adjuncts do take part in UDCs. The treatment
that we propose for adjuncts obviates the need for such an `ad-hoc'
rule.

Clearly many details concerning the syntax of adjuncts are left
untouched here, such as the quite subtle restrictions in word-order
possibilities of certain adjuncts with respect to arguments and with
respect to other adjuncts. In the current framework linguistic
insights concerning these issues could be expressed as constraints on
the resulting subcategorization list (e.g. by means of
LP-constraints).

It should also be stressed that treating adjuncts and arguments on a
par on the level of subcategorization does not imply that observed
differences in the behavior of adjuncts and arguments could not be
handled in the proposed framework. For example the difference of
adjuncts and arguments in the case of left dislocation in Dutch
(exemplified in~\ref{ldis}--\ref{rdis}) can be treated by a lexical
rule that operates on the subcat list before adjuncts are added.
\begin{exam}
\label{ldis}
\begin{flushleft}
De voorstelling duurt een uur\\
The show takes an hour
\end{flushleft}
\end{exam}
\begin{exam}
Een uur, dat duurt de voorstelling
\end{exam}
\begin{exam}
\begin{flushleft}
Arie en Bob wandelen een uur\\
Arie and Bob walk an hour
\end{flushleft}
\end{exam}
\begin{exam}
\label{rdis}
* Een uur, dat wandelen Arie en Bob
\end{exam}

\section{Processing Lexical Rules}

\subsection{Lexical Rules as Constraints on Lexical Categories}

Rather than formalizing the `add-adjuncts' rule as a lexical rule we
propose to use recursive constraints on lexical categories. Such
lexical constraints are then processed using delayed evaluation
techniques.  \footnote{Refer to
\cite{carpenter-lexical} for a proof of Turing equivalence
of simple categorial grammar with recursive lexical rules.}

Such an approach is more promising than an off-line approach that
precomputes the effect of lexical rules by compilation of the lexicon,
as it is unclear how recursive lexical rules can be treated in such an
architecture (especially since some recursive rules can easily lead to
an infinite number of lexical entries, e.g. the adjuncts rule).

Another alternative is to consider lexical rules as `ordinary'
unary rules. If this technique is applied for the lexical rules we
have envisaged here, then (unary) derivations with unbounded length
have to be considered.

If we formalize lexical rules as (complex) constraints on lexical
categories then we are able to use delayed evaluation techniques for
such constraints.

Assume that the `underlying' feature
structure of a verb is given by a definition of `stem' (e.g. as the
example of `wil' above, or as the example of a simple transitive verb
such as `kussen' (to-kiss) in figure~\ref{kus}).

\begin{figure}\begin{center}\small$
\avm{\mbox{\sc verbal}\\
      \mbox{\it sc}: \langle\avm{\mbox{\sc noun}\\\mbox{\it
sem}:\mbox{A}_2}\rangle\\
    \mbox{\it subj}: \avm{\mbox{\sc noun}\\ \mbox{\it sem}: \mbox{A}_1}\\
     \mbox{\it sem$|$nuc$|$qfsoa} : \avm{\mbox{\sc kiss-soa}\\
                          \mbox{\it kisser}: \mbox{A}_1\\
                          \mbox{\it kissed}: \mbox{A}_2}}
$\end{center}\caption{\label{kus}Category for `kussen' (to kiss)}\end{figure}

Such a feature-structure is not the actual category of the verb ---
rather this category is defined with complex constraints with respect
to this base form.  Here the constraint that adds adjuncts to the
subcat list has our special attention, but there is also a constraint
that adds a subject to the subcat list (as part of the inflection
constraint for finite verbs) and a constraint that pushes an element
from the subcat list to slash (to treat unbounded dependencies along
the lines of chapter 9 of \cite{hpsg2}), etc. Thus a lexical
entry might be defined as in figure~\ref{len}.

\begin{figure}\small
\begin{flushleft}
$\mbox{lexical\_entry} (\mbox{A}){\mbox{\tt :-}}$\\
$~~~~ \mbox{stem} (\mbox{B}), ~~~~~~~~~~~\mbox{add\_adj} (\mbox{B},\mbox{C})
 , $\\
$~~~~ \mbox{inflection} (\mbox{C},\mbox{D}), ~\mbox{push\_slash}
(\mbox{D},\mbox{A})  . $\\
\end{flushleft}

\begin{flushleft}$
 \mbox{inflection} (\avm{
 \mbox{\sc verbal}  \\
 \mbox{\it phon} : \mbox{P}\\
 \mbox{\it sc} :\mbox{Sc}\\
 \mbox{\it subj}: \mbox{Subj}}, \avm{
 \mbox{\sc finite}  \\
 \mbox{\it phon} : \mbox{P} \oplus \mbox{``t''} \\
 \mbox{\it sc} :\mbox{Sc}\cdot\langle\mbox{Subj}\rangle \\
 \mbox{\it subj}: \mbox{Subj}}).$\\
\end{flushleft}
\caption{\label{len}
A lexical entry is defined with respect to a base form using complex
constraints. Subject addition is a
constraint associated with finite inflection. }
\end{figure}

Lexical rules are regarded as (complex) constraints in this framework
because it allows an implementation using delayed evaluation
techniques from logic programming.
The idea is that a certain
constraint is only (partially) evaluated if `enough' information is
available to do so successfully. As a relatively simple example we consider the
constraint that is responsible for adding a subject as the last
element on a subcat list of finite verbs. As a lexical rule we might
define:
\begin{center}\small
$
\avm{\mbox{\sc finite}\\
      \mbox{\it subj}: \mbox{Subj}\\
      \mbox{\it sc}: \mbox{Sc}}
\Longrightarrow
\avm{\mbox{\it sc}: \mbox{Sc}\cdot\langle\mbox{Subj}\rangle}
$
\end{center}
If we use constraints the definition can be given as in figure~\ref{len},
as part of the constraint associated with finite morphology. Note that
the two approaches are not equivalent. If we use lexical rules then we
have to make sure that the add-subject rule should be applied only
once, and only for finite verbs. As a constraint we simply call the
constraint once at the appropriate position.

The concatenation constraint (associated with the `dot' notation) is
defined as usual:
\begin{flushleft}
$
 \mbox{concat} (\langle\rangle,\mbox{A},\mbox{A}).$\\
$
 \mbox{concat} (\langle \mbox{B} | \mbox{C}\rangle ,
\mbox{A},\langle \mbox{B} | \mbox{D}\rangle ){\mbox{\tt :-}}$\\
$
{}~~~~ \mbox{concat} (\mbox{C},\mbox{A},\mbox{D}). $\\
\end{flushleft}
If this constraint applies on a category of which the subcat list is
not yet fully specified (for example because we do not yet know how
many adjuncts have been added to this list) then we cannot yet compute
the resulting subcat list. The constraint can be successfully applied
if either one of the subcat lists is instantiated: then we obtain a
finite number of possible solutions to the constraint.\\

The relation {\it add\_adj} recursively descends through a
subcategorization list and at each position either adds or does not
add an adjunct (of the appropriate type). Its definition is given in
figure~\ref{plad}.
\begin{figure}
\begin{flushleft}\small
$
 \mbox{add\_adj} (\avm{
 \mbox{\sc sign}  \\
 \mbox{\it sc} :\mbox{A} \\
 \mbox{\it sem} :\mbox{B} \\
 \mbox{\it subj}:\mbox{Subj}},
                  \avm{
 \mbox{\sc sign}  \\
 \mbox{\it sc} :\mbox{J} \\
 \mbox{\it sem} :\mbox{K} \\
 \mbox{\it subj} : \mbox{Subj}}){\mbox{\tt :-}}$\\
$
{}~~~~ \mbox{add\_adj} (\mbox{A},\mbox{J},\mbox{B},\mbox{K}). $\\
\end{flushleft}
\begin{flushleft}\small
$ \mbox{add\_adj} (\langle\rangle,\langle\rangle,\mbox{A},\mbox{A}).   $\\
$ \mbox{add\_adj} (\langle \mbox{C} | \mbox{D}\rangle ,
\langle \mbox{C} | \mbox{E}\rangle ,\mbox{A},\mbox{B}){\mbox{\tt :-}}  $\\
$ ~~~~ \mbox{add\_adj} (\mbox{D},\mbox{E},\mbox{A},\mbox{B}). $\\
$ \mbox{add\_adj} (\mbox{A},\langle \avm{
 \mbox{\sc adverbial}  \\
 \mbox{\it mod} :\avm{
 \mbox{\sc mod}  \\
 \mbox{\it arg} :\mbox{B} \\
 \mbox{\it val} :\mbox{E}}} | \mbox{D}\rangle ,\mbox{B},\mbox{C}){\mbox{\tt
:-}}$\\
$ ~~~~ \mbox{add\_adj} (\mbox{A},\mbox{D},\mbox{E},\mbox{C}).
$\\
\end{flushleft}
\caption{\label{plad}Definite clause specification of `add\_adj' constraint.}
\end{figure}
Note that it is assumed in this definition that the scope of
(operator-type) adverbials is given by the order in which they are put
in in the subcategorization list, i.e. in the obliqueness order.
\footnote{Cf. \cite{kasper-adjuncts} for discussion of this
point, also in relation with adjuncts that introduce quantifiers. Note
that in our approach different possibilities can be defined.}

\subsection{Delayed evaluation}

For our current purposes, the co-routining facilities offered by
Sicstus Prolog are powerful enough to implement a delayed evaluation
strategy for the cases discussed above. For each constraint we declare
the conditions for evaluating a constraint of that type by means of a
{\tt block} declaration.  For example the {\it concat} constraint is
associated with a declaration:\\

\noindent $\mbox{\tt :- block } \mbox{\it concat}(-,?,-).$\\

\noindent
This declaration says that evaluation of a call to {\it
concat} should be delayed if both the first and third arguments are
currently variable (uninstantiated, of type {\sc top}).  It is clear
from the definition of {\it concat} that if these arguments are
instantiated then we can evaluate the constraint in a top-down manner
without risking non-termination. E.g. the goal $\mbox{\it
concat}(\langle\mbox{A},\mbox{B}\rangle,\mbox{C},\mbox{D})$ succeeds
by instantiating D as the list $\langle
\mbox{A},\mbox{B}|\mbox{C}\rangle$.

Note that block declarations apply recursively.  If the third argument
to a call to {\it concat} is instantiated as a list with a variable
tail, then the evaluation of the recursive application of that goal
might be blocked; e.g. evaluation of the goal $\mbox{\it
concat}(\mbox{A},\langle \mbox{Sj}\rangle,\langle\mbox{B}|\mbox{C}\rangle)$
succeeds either with both A and C instantiated as the empty list and by
unifying Sj and B, or with A instantiated as the list
$\langle\mbox{B}|\mbox{D}\rangle$ for which the constraint $\mbox{\it
concat}(\mbox{D},\langle\mbox{Sj}\rangle,\mbox{C})$ has to be satisfied.
Similarly, for each of the other constraints we declare the
conditions under which the constraint can be evaluated. For the
{\it add\_adj} constraint we define:\\

\noindent $\mbox{\tt :- block } \mbox{\it add\_adj}(?,-,?,?).$\\

\noindent
One may wonder whether in such an architecture enough information will
ever become available to allow the evaluation of any of the
constraints. In general such a problem may surface: the parser then
finishes a derivation with a large collection of constraints that it
is not allowed to evaluate --- and hence it is not clear whether the
sentence associated with that derivation is in fact grammatical (as
there may be no solutions to these constraints).

The strategy we have used successfully so-far is to use the structure
hypothesized by the parser as a `generator' of information. For
example, given that the parser hypothesizes the application of rules,
and hence of certain instantiations of the subcat list of the
(lexical) head of such rules, this provides information on the
subcat-list of lexical categories. Keeping in mind the definition of a
lexical entry as in figure~\ref{len} we then are able to evaluate each
of the constraints on the value of the subcat list in turn, starting
with the {\it push\_slash} constraint, up through the {\it inflection}
and {\it add\_adj} constraints. Thus rather than using the constraints
as `builders' of subcat-lists the constraints are evaluated by
checking whether a subcat-list hypothesized by the parser can be
related to a subcat-list provided by a verb-stem. In other words, the
flow of information in the definition of {\it lexical\_entry} is not
as the order of constraints might suggest (from top to bottom) but
rather the other way around (from bottom to top).

\section{Final remarks}

We illustrated that recursive lexical constraints might be useful from a
linguistic perspective. If lexical rules are formalized as complex
constraints on lexical categories then methods from logic programming
can be used to implement such constraints.

Note that complex constraints and delayed evaluation techniques are
also useful in other areas of linguistic desciption.  For example we used
the same methods to define and process HPSG's {\sc foot feature principle}.
The method may also be applied to implement HPSG's binding theory.

As a testcase we improved upon the HPSG analysis of (Germanic) verb
clusters and adjuncts by treating adjuncts as categories that are on
the subcat list by virtue of a complex constraint. The fragment that
has been implemented with the methods described is much larger than
the discussion in the previous sections suggest, but includes
treatments of {\em extraposition, ipp, modal inversion, participium
inversion, the third construction, partial-vp topicalisation, particle
verbs, verb-second, subject raising, subject control,
raising-to-object, object control and clitic climbing} in Dutch.

\end{document}